\documentclass[letter, 10pt, conference]{ieeeconf}      

\IEEEoverridecommandlockouts                              
\usepackage[utf8]{inputenc}
\usepackage[T1]{fontenc}
\usepackage{times,multirow,float}
\usepackage{graphicx}
\usepackage{epsfig,xspace,layout}
\usepackage{color}
\usepackage{amsfonts}
\usepackage{times}
\usepackage{amssymb}
\usepackage{amsmath}
\usepackage{hyperref}
\usepackage{amsmath,bm}
\usepackage{rotating}
\usepackage{mathrsfs}
\usepackage{mathtools}
\usepackage{makeidx}
\usepackage[]{threeparttable}
\usepackage{dsfont}
\usepackage{cite}
\usepackage{xcolor}
\usepackage{cleveref}

\usepackage[ruled, lined, linesnumbered, commentsnumbered, longend]{algorithm2e}
\usepackage{algpseudocode}
\usepackage{tikz}
\usetikzlibrary{shapes,arrows,positioning,calc}
\usepackage{siunitx}
\usepackage{multicol,lipsum}
\usepackage{subcaption}
\usepackage[font=footnotesize]{caption}
\usepackage{url}
\usepackage{booktabs}
\usepackage{lineno}
\usepackage{hhline}
\usepackage[Symbol]{upgreek}
\usepackage{float}

\SetKwComment{Comment}{$\triangleright$\ }{}
\makeatletter
\renewcommand{\Indentp}[1]{%
  \advance\leftskip by #1
  \advance\skiptext by -#1
  \advance\skiprule by #1}%
\renewcommand{\Indp}{\algocf@adjustskipindent\Indentp{\algoskipindent}}
\renewcommand{\Indm}{\algocf@adjustskipindent\Indentp{-\algoskipindent}}
\makeatother

\title{\LARGE \bf
Trust-Region Neural Moving Horizon Estimation for Robots
}

\author{Bingheng Wang, Xuyang Chen, and Lin Zhao
\thanks{
\setlength{\fboxrule}{0.4pt}\setlength{\fboxsep}{6pt}%
  \fbox{\parbox{0.92\columnwidth}{
    Accepted for publication in the Proceedings of the 2024 IEEE
    International Conference on Robotics and Automation (ICRA).
    \copyright~2024 IEEE. Personal use of this material is permitted.
    Permission from IEEE must be obtained for all other uses, in any
    current or future media, including reprinting/republishing this
    material for advertising or promotional purposes, creating new
    collective works, for resale or redistribution to servers or
    lists, or reuse of any copyrighted component of this work in
    other works. DOI: 10.1109/ICRA57147.2024.10611703}}
This work was supported by the Singapore Ministry of Education Tier 1 Academic Research Fund (22-5460-A0001) and Tier 2 AcRF (T2EP20123-0037). (Corresponding author: Lin Zhao.) }
\thanks{These authors are with the Department of Electrical and Computer Engineering, National University of Singapore, 4 Engineering Drive 3, 117583 Singapore, Singapore
        {\tt\small $\left\{ {} \right.$wangbingheng, chenxuyang$\left.\right\}$@u.nus.edu}, {\tt\small elezhli@nus.edu.sg}}
\thanks{The source code of this work is available at \url{https://github.com/BinghengNUS/TR-NeuroMHE} }
}

\begin{document}

\maketitle
\thispagestyle{empty}
\pagestyle{empty}

\begin{abstract}
Accurate disturbance estimation is essential for safe robot operations. The recently proposed neural moving horizon estimation (NeuroMHE), which uses a portable neural network to model the MHE's weightings, has shown promise in further pushing the accuracy and efficiency boundary. Currently, NeuroMHE is trained through gradient descent, with its gradient computed recursively using a Kalman filter. This paper proposes a trust-region policy optimization method for training NeuroMHE. We achieve this by providing the second-order derivatives of MHE, referred to as the MHE Hessian. Remarkably, we show that many of the intermediate results used to obtain the gradient, especially the Kalman filter, can be efficiently reused to compute the MHE Hessian. This offers linear computational complexity with respect to the MHE horizon. As a case study, we evaluate the proposed trust region NeuroMHE on real quadrotor flight data for disturbance estimation. Our approach demonstrates highly efficient training in under $5\ {\rm min}$ using only $100$ data points. It outperforms a state-of-the-art neural estimator by up to $68.1\%$ in force estimation accuracy, utilizing only $1.4\%$ of its network parameters. Furthermore, our method showcases enhanced robustness to network initialization compared to the gradient descent counterpart.

\end{abstract}


\section{Introduction}
Disturbances are widespread in robot operations, arising from various sources like system uncertainties~\cite{corradini2001robust}, aerodynamic influences~\cite{powers2013influence}, ground friction forces~\cite{hashemi2016integrated}, and hydrodynamic disturbances~\cite{chu2022path}. Addressing these disturbances is vital in robotic control to prevent significant performance degradation~\cite{coelho2005path,park2008adaptive}. However, developing a versatile model to capture these disturbances across environments is typically impractical due to their complexity. Therefore, online precise disturbance estimation that adapts to environments is crucial for safe and effective robot operations.

Existing works in disturbance estimation mainly utilize either model-based~\cite{yuksel2014nonlinear,tomic2014unified,huang2018disturbance,hentzen2019disturbance,mckinnon2020estimating} or model-free~\cite{liu2018neuro,bauersfeld2021neurobem,shi2021neural} methods. While model-based estimators are data-efficient, they generally have limited dynamic response range, since the models used are typically limited to specific slowly-varying disturbances~\cite{yuksel2014nonlinear,tomic2014unified,huang2018disturbance}. Another challenge lies in their heavy reliance on manually tuning numerous parameters~\cite{hentzen2019disturbance,mckinnon2020estimating}, which demands significant experimental efforts and expert knowledge. In contrast, model-free approaches using neural networks (NN) can achieve high performance across various disturbance scenarios with minimal expert knowledge. Examples include estimating system uncertainties for robotic joints~\cite{liu2018neuro} and aerodynamic effects for aerial robots~\cite{bauersfeld2021neurobem,shi2021neural}. In practice, these NN-based estimators typically use large network models for training, along with substantial disturbance data and complicated learning curricula.

\begin{figure}[t!]
	\centering
	{\includegraphics[width=0.975\columnwidth]{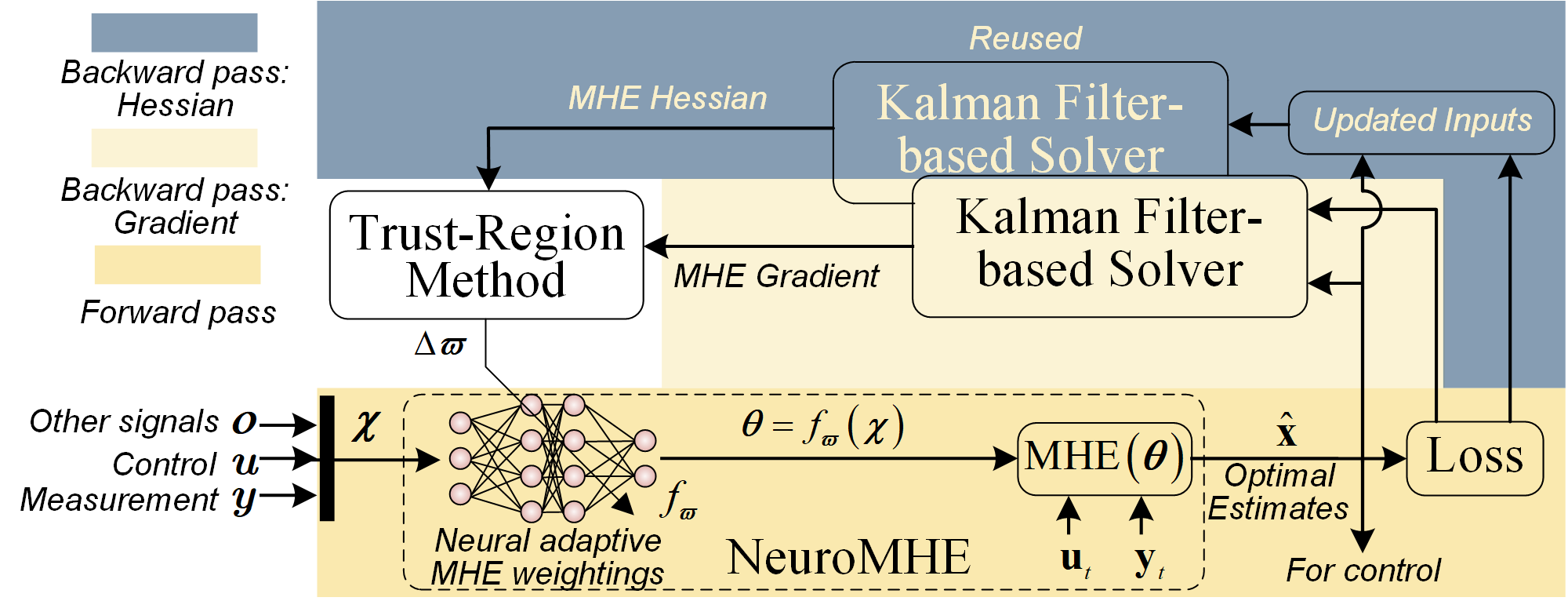}}
	\caption{\footnotesize Learning pipelines of the trust-region NeuroMHE. Currently, NeuroMHE is trained via gradient descent with its gradient computed recursively using a Kalman filter. This paper enhances NeuroMHE training with the second-order trust-region method. Interestingly, we show that the MHE Hessian can be obtained recursively using the same Kalman filter with just minor modifications to its inputs.}
\label{fig:neuromhe learning pipeline}	
\end{figure}

To leverage advantages from both model-free and model-based methods, we recently proposed the neural moving horizon estimation (NeuroMHE)~\cite{10313083}. This algorithm fuses a portable neural network into an MHE to realize accurate estimation and fast online adaptation (See the dashed block in Fig.~\ref{fig:neuromhe learning pipeline}). MHE is an optimal state estimator using control theory for dynamic nonlinear optimization online, with a receding horizon strategy. It has consistently demonstrated superior performance in disturbance estimation for both ground~\cite{kraus2013moving} and aerial~\cite{wang2021differentiable,papadimitriou2022external} robots. Nevertheless, the parameter tuning remains a challenge for vanilla MHE, especially when dealing with dynamic and nonlinear parameters~\cite{robertson1996moving}. In contrast, NeuroMHE tackles this challenge by modeling its parameters using the neural network trained through gradient descent. The gradient includes the derivative of MHE's optimal solution trajectory with respect to (w.r.t) the parameters, computed recursively using a Kalman filter. This approach results in linear computational complexity relative to the MHE horizon, significantly improving training efficiency over state-of-the-art auto-tuning MHE methods~\cite{esfahani2021reinforcement,muntwiler2022learning}.

In this paper, we enhance NeuroMHE training by providing second-order derivative information of MHE (i.e., the MHE Hessian). This allows us to employ the trust-region method (TRM), a second-order optimization technique, for training NeuroMHE. The trust-region method offers adaptive step-size updates, faster convergence, and improved robustness to initialization. It has shown efficacy across machine learning and robotics, including policy optimization~\cite{schulman2015trust} and, more recently, synergizing optimal control with inverse reinforcement learning~\cite{cao2023trust}. Fig.~\ref{fig:neuromhe learning pipeline} outlines the learning pipelines of the trust-region NeuroMHE. Two key ingredients of our algorithm are the gradient and the Hessian trajectories of the MHE's optimal solutions w.r.t the tuning parameters in the backward passes. By differentiating the Karush-Kuhn-Tucker (KKT) conditions of the MHE optimization problem in the forward pass, we can efficiently obtain the MHE gradient trajectory using a Kalman filter, as developed in~\cite{10313083}. Interestingly, we further show that the same Kalman filter can be reused to compute the MHE Hessian trajectory with only minor modifications to its inputs, which are acquired through double differentiation of the KKT conditions. This preserves the linear computational complexity w.r.t the MHE horizon, further ensuring the scalability of our method to high-dimension estimation problem.

We evaluate the proposed trust-region NeuroMHE in estimating complex aerodynamic disturbances using open-source real quadrotor flight data as a case study. Compared to the state-of-the-art estimator NeuroBEM~\cite{bauersfeld2021neurobem}, our approach achieves: 1) highly efficient training less than $5\ {\rm min}$ with only $100$ data points; 2) using only $1.4\%$ of the NeuroBEM's network parameters; and 3) reducing the overall force estimation error by up to $68.1\%$. Further comparisons with the gradient descent counterpart show that, although computing the MHE Hessian trajectory naturally takes longer than the gradient trajectory, our method significantly reduces both the total training episodes and the overall training time. Additionally, our approach exhibits strong robustness to the initialization of the neural network. In summary, our contributions are threefold:
\begin{enumerate}
    \item We propose a trust-region policy optimization method for training the recently developed NeuroMHE. 
    \item We show that much of the computation used to obtain the MHE gradient trajectory, particularly the Kalman filter, can be efficiently reused for computing the MHE Hessian trajectory in a recursive manner.
    \item We validate the effectiveness of our approach using a real flight dataset, showing highly efficient training and superior performance compared to the state-of-the-art method in aerodynamic force estimation accuracy.
\end{enumerate}

The rest of this paper is organized as follows. Section~\ref{sec:problem} briefly reviews the formulation of NeuroMHE. In Section~\ref{sec:Hessian}, we detail the process of reusing the Kalman filter to compute the MHE Hessian. Section~\ref{sec:Trust Region} develops the proposed trust-region NeuroMHE training method. Simulation results on the real dataset are reported in Section~\ref{sec:experiment}. We conclude this paper and discuss our future work in Section~\ref{sec:conclusion}.

\section{Preliminary of NeuroMHE}\label{sec:problem}

\subsection{Moving Horizon Estimation}
MHE is a model-based optimal state estimator. Without loss of generality, we assume that the robotic dynamics model used in MHE takes the following form:
\begin{equation}
\dot{\bm{x}} =\bm{f}\left ( \bm{x},\bm{u},\bm{w}  \right ),\
\bm{y}=\bm{h}\left ( \bm{x} \right ) + \boldsymbol\nu ,
\label{eq:model}
\end{equation}
where $\bm{x} \in \mathbb{R}^n$ is the system state, $\bm{u} \in \mathbb{R}^m$ is the control input, $\bm{f}$ is the system model, $\bm{w} \in \mathbb{R}^w$ is the process noise, $\bm{h}$ is the output function, and $\bm{y} \in \mathbb{R}^l$ denotes the measurement affected by the noise $\boldsymbol\nu \in \mathbb{R}^l$. At time $t$, given the most recent measurements $\mathbf{y}_t=\left \{ \bm{y}_{k} \right \}_{k=t-N}^{t}$ and control inputs $\mathbf{u}_t=\left \{ \bm{u}_{k} \right \}_{k=t-N}^{t-1}$ collected in a moving data window of horizon $N$, the MHE estimator optimizes over $\mathbf{x}=\left \{ \bm{x}_{k} \right \}_{k=t-N}^{t}$ and $\mathbf{w}=\left \{ \bm{w}_{k} \right \}_{k=t-N}^{t-1}$ at each time step $t\geq N$ by solving the following nonlinear optimization problem online.
\begin{subequations}
\begin{align}
\begin{split}
\mathop {\min }\limits_{\bm {\mathop{ {\rm x}}\nolimits}, \bm {\mathop{ {\rm w}}\nolimits} } J & = \underbrace{\frac{1}{2}\left\| {{{ {\bm x}}_{t - N}} - {{\hat{\bm x}}_{t - N}}} \right\|_{\bm P}^2}_{\rm arrival \ cost}\\
 &\quad + \underbrace{\frac{1}{2}\sum\limits_{k = t - N}^t {\left\| {{\bm y_k} - {\bm h}\left( {{{ {\bm x}}_{k}}} \right)} \right\|_{{\bm R_k}}^2}  + \frac{1}{2}\sum\limits_{k = t - N}^{t - 1} {\left\| {{{\bm w} _k}} \right\|_{{{\bm Q}_k}}^2}}_{\rm running \ cost} 
\end{split}
\label{eq:mhe cost}
\\
    {\rm s.t.}\ & { {\bm x}_{k + 1}} = {\mathbf {f}}\left( {{{ {\bm x}}_{k}},{{\bm u}_k},{{{\bm w}} _k},\Delta t} \right),
\label{eq:mhe equality constraint}
\end{align}
\label{eq:mhe}%
\end{subequations}
where $\bm P \in \mathbb{R}^{n\times n}$, ${\bm R}_k \in \mathbb{R}^{l\times l}$, and ${\bm Q}_k \in \mathbb{R}^{w\times w}$ are the positive-definite weighting matrices, ${\mathbf f}$ is the discrete-time model of $\bm {f}$ with the step-size of $\Delta t$ for predicting the state, and ${{ \hat{\bm x}}_{t - N}}$ is the filter priori, chosen as the MHE's optimal estimate ${{\hat {\bm x}}_{t - N\left| {t-1} \right.}}$ of $\bm x_{t-N}$ obtained at $t-1$~\cite{alessandri2010advances}. We denote by $\hat {\mathbf{x}} = \left\{ {{{\hat {\bm x}}_{k\left| t \right.}}} \right\}_{k = t - N}^t$ and $\hat {\mathbf{w}} = \left\{ {{{\hat {\bm w}}_{k\left| t \right.}}} \right\}_{k = t - N}^{t - 1}$ the MHE's optimal solutions to Problem (\ref{eq:mhe}) at time step $t$.

\subsection{Formulation of NeuroMHE}
The weighting matrices in the cost function (\ref{eq:mhe cost}) are the tuning parameters, which determine the MHE performance. For ease of presentation, we collect them in a vector $\boldsymbol\theta =\left [ vec\left ( {\bm P} \right ),\left \{ vec\left ( \bm {R}_k \right ) \right \}^{t}_{k=t-N},\left \{  vec\left ( \bm {Q}_k \right )\right \}^{t-1}_{k=t-N} \right ] \in \mathbb{R}^{p}$ where $vec\left(  \cdot  \right)$ denotes the vectorization of a given matrix. Tuning $\boldsymbol\theta$ typically necessitates expert knowledge of the covariances of the noises entering robotic systems. In practice, identifying these noise covariances is difficult due to their large number and strong coupling. It becomes even more demanding when these values exhibit dynamic behavior. 

NeuroMHE tackles the above challenge by modeling $\boldsymbol\theta$ with a portable neural network:
\begin{equation}
    \bm \theta  = {{\bm f}_{\boldsymbol \varpi} }\left( {\boldsymbol \chi}  \right).
    \label{eq:nn parameterization}
\end{equation}
The neural network can accept various signals as its inputs $\boldsymbol \chi$, including measurements $\bm y$, controls $\bm u$, and other external signals $\bm o$, depending on specific applications. Its parameters $\boldsymbol \varpi$ are trained using powerful machine learning techniques. Here, $\boldsymbol \varpi$ represents the tuning parameters for NeuroMHE, and we parameterize Problem (\ref{eq:mhe}) as $\rm{NeuroMHE}\left(\boldsymbol \varpi \right)$, with the MHE's solution denoted as ${\hat {\mathbf{x}}\left(\boldsymbol \varpi \right)}$. To tune MHE, a high-level optimization problem can be formulated as follows:
\begin{subequations}
\begin{align}
\mathop {\min }\limits_{\boldsymbol \varpi}\   &L\left( {\hat {\mathbf{x}}\left(\boldsymbol \varpi \right)} \right)
\\
{\rm s.t.}\ & \hat {\mathbf{x}}\left(\boldsymbol \varpi \right)\ {\rm generated \ by} \ {\rm{NeuroMHE}}\left(\boldsymbol \varpi \right).
\label{eq:low level neuromhe}
\end{align}
\label{eq:bi level neuromhe}%
\end{subequations}
where $L\left( {\hat {\mathbf{x}}\left(\boldsymbol \varpi \right)} \right)$ is a loss function assessing NeuroMHE's performance. It can be built upon estimation errors when ground truth disturbance data is available. Alternatively, it can be tailored to penalize tracking errors in applications involving robust trajectory tracking control.

\section{Analytical Hessian Trajectory}\label{sec:Hessian}
Our previous work~\cite{10313083} solved Problem (\ref{eq:bi level neuromhe}) using gradient descent, with the gradient computed recursively through a Kalman filter. In this section, our objective is to efficiently reuse the same Kalman filter and much of the computation initially employed to obtain the gradient. We intend to apply these resources in computing the MHE Hessian for training NeuroMHE with second-order optimization techniques. 

\subsection{MHE Gradient}
\label{subsec: kalman gradient}
The gradient of $L\left( {\hat {\mathbf{x}}\left(\boldsymbol \varpi \right)} \right)$ w.r.t $\boldsymbol \varpi$ can be computed using the chain rule:
\begin{equation}
     \nabla_{\boldsymbol \varpi} L\left( {\hat {\mathbf{x}}}\left(\boldsymbol \varpi \right) \right)  = \nabla_{\hat{\mathbf{x}}} L \left( {\hat {\mathbf{x}}} \right) \nabla_{\boldsymbol \theta}\hat {\mathbf{x}}\left( \boldsymbol \theta \right) \nabla_{\boldsymbol \varpi} {\boldsymbol \theta\left(\bm \varpi \right)}.
    \label{eq: gradient of loss}
\end{equation}
The gradients $\nabla_{\hat{\mathbf{x}}} L$ and $\nabla_{\boldsymbol \varpi} {\boldsymbol \theta}$ are straightforward to obtain as both $L\left( {\hat {\mathbf{x}}} \right)$ and $\boldsymbol \theta\left(\boldsymbol \varpi \right)$ are explicit functions. The main challenge lies in the computation of $\nabla_{\boldsymbol \theta}\hat {\mathbf{x}}$ (the MHE gradient). Recall that $\hat{\mathbf{x}}\left(\boldsymbol{\theta}\right)$ is the MHE's solution, with its dependence on $\boldsymbol{\theta}$ implicitly defined via the KKT conditions. This implicit definition enables us to compute $\nabla_{\boldsymbol \theta}\hat {\mathbf{x}}$ by differentiating through the KKT conditions of Problem (\ref{eq:mhe}). To proceed, we associate the equality constraints~(\ref{eq:mhe equality constraint}) with the dual variables $\bm \lambda = \left\{ {{{ {\bm \lambda}}_{k}}} \right\}_{k = t - N}^{t-1}$ and denote $\bm \lambda^* \in \mathbb{R}^{n}$ as their optimal values. The corresponding Lagrangian can be written as
\begin{equation}
    {\cal L} = J+ \sum\limits_{k = t - N}^{t - 1} {{\bm \lambda} _k^T\left( {{{ {\bm x}}_{k + 1}} - {\mathbf f}\left( {{{ {\bm x}}_{k}},{{\bm u}_k},{{\bm w} _{k}},\Delta t} \right)} \right)}.
    \label{eq: lagrangian}
\end{equation}
Then the KKT conditions of Problem (\ref{eq:mhe}) at ${\hat{\bm {\mathop{ {\rm x}}\nolimits}} }$, ${\hat{\bm {\mathop{ {\rm w}}\nolimits}} }$, and $\bm \lambda^*$ are given by
\begin{subequations}
\allowdisplaybreaks
\begin{align}
\begin{split}
\nabla_{{\hat {\bm x}}_{t-N|t}}{\cal L}\left ( {\hat {\bm x}}_{t-N|t},{\hat {\bm x}}_{t-N},\hat{\bm w}_{t-N|t},\boldsymbol{\lambda}_{t-N}^{*},\boldsymbol\theta \right )&=\bm 0, 
\end{split}
\label{eq:kkt boundary}\\
\begin{split}
\nabla_{{\hat {\bm x}}_{k|t}}{\cal L}\left ( {\hat {\bm x}}_{k|t},{\hat {\bm w}}_{k|t},\boldsymbol\lambda_{k}^{*}, \boldsymbol\lambda_{k-1}^{*},\boldsymbol\theta \right )&= \bm 0,
\end{split}
\label{eq:kkt x}\\
\begin{split}
\nabla_{{\hat {\bm w}}_{k|t}}{\cal L}\left ( {\hat {\bm x}}_{k|t},{\hat {\bm w}}_{k|t},\boldsymbol\lambda_{k}^{*},\boldsymbol\theta \right )& =\bm 0, 
\end{split}
\label{eq:kkt noise}\\
\begin{split}
{\nabla _{\lambda^{*}_k} }{{\cal { L}}}  =  {{\hat {\bm x}}_{k + 1\left| t \right.}}- {\mathbf f}\left( {{{\hat {\bm x}}_{k\left| t \right.}},{\bm u_k},{\hat{\bm w} _{k\left| t \right.}},\Delta t} \right)  & = \bm 0,
\end{split}
\label{eq:kkt lambda}
\end{align}
\label{eq:kkt}%
\end{subequations}
Differentiating the KKT conditions (\ref{eq:kkt}) w.r.t $\boldsymbol{\theta}$ yields the following differential KKT conditions.
\begin{subequations}
\begin{align}
    \begin{split}
\frac{{d{\nabla _{{\hat {x}_{t - N|t}}}}{\cal L}}}{{d\boldsymbol \theta }} & =  { {\bm L}_{t - N}^{xx}} \nabla_{\boldsymbol \theta} {\hat{\bm x}_{t-N|t}} \mathrlap{- {\bm P}\nabla_{\boldsymbol \theta}{\hat{\bm x}_{t-N}}+ {\bm L}_{t - N}^{x\theta }} \\
& \quad  + \mathrlap{{\bm L}_{t - N}^{xw}\nabla_{\boldsymbol \theta} {\hat{\bm w}_{t-N|t}}- {\bm F}_{t - N}^T\nabla_{\boldsymbol \theta} {{\boldsymbol \lambda}_{t-N}^*}= {\bm 0},}
    \end{split}
    \label{eq:boundary differential KKT}\\
    \begin{split}
\frac{{d{\nabla _{{\hat{x}_k|t}}}{\cal { L}}}}{{d\boldsymbol \theta }} & = {\bm L}_k^{xx}\nabla_{\boldsymbol \theta} {\hat{\bm x}_{k|t}} + {\bm L}_k^{xw }\nabla_{\boldsymbol \theta} {\hat{\bm w}_{k|t}} - \mathrlap{{\bm F}_k^T\nabla_{\boldsymbol \theta} {{\boldsymbol \lambda}_{k}^*}}\\
&\quad + \nabla_{\boldsymbol \theta} {{\boldsymbol \lambda}_{k-1}^*} + {\bm L}_k^{x\theta } = {\bm 0},
    \end{split}
    \label{eq:x differential KKT}\\
    \begin{split}
\frac{{d{\nabla _{{\hat{w} _{k|t}}}}{\cal { L}}}}{{d\boldsymbol \theta }} & = {\bm L}_k^{w x}\nabla_{\boldsymbol \theta} {\hat{\bm x}_{k|t}} + {\bm L}_k^{w w}\nabla_{\boldsymbol \theta} {\hat{\bm w}_{k|t}} \\
&\quad - {\bm G}_k^T\nabla_{\boldsymbol \theta} {{\boldsymbol \lambda}_{k}^*} + {\bm L}_k^{w \theta } = {\bm 0},
    \end{split}
    \label{eq:w differential KKT}\\
    \begin{split}
\frac{{d{\nabla _{{\lambda^{*} _k}}}{\cal { L}}}}{{d\boldsymbol \theta }} & = \nabla_{\boldsymbol \theta} {\hat{\bm x}_{k+1|t}} - {\bm F_k} \nabla_{\boldsymbol \theta} {\hat{\bm x}_{k|t}} - {\bm G_k} \nabla_{\boldsymbol \theta} {\hat{\bm w}_{k|t}} = {\bm 0},
    \end{split}
    \label{eq:costate differential KKT}
\end{align}
\label{eq:differential KKT}%
\end{subequations}
where ${\bm F_k} = \nabla_{{\hat {\bm x}}_{k\left| t \right.}} \mathbf {f}$, ${\bm G_k} = \nabla_{\hat{\bm w} _{k\left| t \right.}} \mathbf{f}$, ${\bm H_k} = \nabla_{{\hat {\bm x}}_{k\left| t \right.}} \bm h$, $\bm \lambda^*_t = \bm 0$, ${\bm L}_k^{xx}  = \frac{{{\partial ^2}{\cal L}}}{{\partial \hat {\bm x}_{k\left| t \right.}^2}}$, ${\bm L}_k^{xw }  = \frac{{{\partial ^2} {\cal L}}}{{\partial {{\hat {\bm x}}_{k\left| t \right.}}\partial {\hat{\bm w} _{k\left| t \right.}}}}$, ${\bm L}_k^{x\theta } = \frac{{{\partial ^2}{\cal L}}}{{\partial {{\hat {\bm x}}_{k\left| t \right.}}\partial {\bm \theta} }}$, ${\bm L}_k^{w w }  = \frac{{{\partial ^2} {\cal L}}}{{\partial {\hat{\bm w}} _{k\left| t \right.}^2}}$, ${\bm L}_k^{w x}  = \frac{{{\partial ^2} {\cal L}}}{{\partial {{\hat{\bm w}} _{k\left| t \right.}}\partial {{\hat {\bm x}}_{k\left| t \right.}}}}$, and ${\bm L}_k^{w \theta }  = \frac{{{\partial ^2}{\cal L}}}{{\partial {\hat{\bm w} _{k\left| t \right.}}\partial {\bm \theta} }}$. 

The above differential KKT conditions (\ref{eq:differential KKT}) can be easily shown to be the same as  the KKT conditions of the following linear MHE problem:
\begin{subequations}
\begin{align}
\begin{split}
\mathop {\min }\limits_{\bm {\mathop{\rm X}\nolimits},\bm {\mathop{\rm W}\nolimits}} {J_2} & = \frac{1}{2}{\rm{Tr}}\left\| {{{ {\bm X}}_{t - N}} - \hat{\bm X}_{t-N}} \right\|_{\bm P}^2\\
&\quad + {\rm{Tr}}\sum\limits_{k = t - N}^t {\left( {\frac{1}{2} {\bm X}_{k}^T\bar {\bm L}_k^{xx}{{ {\bm X}}_{k}} + {{\bm W}_k}^T {\bm L}_k^{w x}  {\bm X}_{k} } \right)} \\
&\quad + {\rm{Tr}}\sum\limits_{k = t - N}^{t - 1} {\left( {\frac{1}{2}{\bm W}_k^T {\bm L}_k^{w w}{{\bm W}_k} +{{{\left( {{\bm L}_k^{w \theta }} \right)}^T}{{\bm W}_k}} }\right)}\\
&\quad + {\rm{Tr}}\sum\limits_{k = t - N}^{t} {\left( {{\left( {{\bm L}_k^{x\theta }} \right)}^T} {{ {\bm X}}_{k}}  \right)}  
\end{split}
\label{eq:auxiliary mhe cost}\\
{\rm s.t.}\ & {{ {\bm X}}_{k + 1}} = {\bm F_k}{{ {\bm X}}_{k}} + {\bm G_k}{{\bm W}_k},
\label{eq:auxiliary dynamics}
\end{align}
\label{eq: auxiliary MHE}%
\end{subequations}
where ${\mathbf{X}} = \left\{ {{{ {\bm X}}_{k}}} \right\}_{k = t - N}^t$, ${\mathbf{W}} = \left\{ {{{ {\bm W}}_{k}}} \right\}_{k=t - N}^{t - 1}$, $\bar {\bm L}_k^{xx} = \frac{{{\partial ^2}{\cal {\bar L}}}}{{\partial \hat {\bm x}_{k\left| t \right.}^2}}$ with ${\cal {\bar L}}$ obtained by excluding the arrival cost (See (\ref{eq:mhe cost})) from ${\cal L}$, and ${\rm Tr}\left(  \cdot  \right)$ is the matrix trace. Hence, the optimal solution $\hat{\mathbf{X}}$ to Problem (\ref{eq: auxiliary MHE}) is exactly the gradient $\nabla_{\boldsymbol \theta}\hat {\mathbf{x}}$, denoted as $\hat{\mathbf{X}}=\left\{ {{{ \hat{\bm X}}_{k}}} \right\}_{k = t - N}^t=\left \{ \nabla_{\boldsymbol \theta} {\hat{\bm x}_{k|t}} \right \}_{k=t-N}^{t}$. 

Note that (\ref{eq: auxiliary MHE}) has a quadratic cost function (\ref{eq:auxiliary mhe cost}) with linear dynamics constraints (\ref{eq:auxiliary dynamics}). It is possible to obtain $\hat{\mathbf{X}}$ analytically. In~\cite{10313083}, we proposed a recursive method for computing $\hat{\mathbf{X}}$ using a Kalman filter. The details of this method will be presented in Algorithm~\ref{alg: analytical solution} (See \ref{subsec: kalman solver}). Next, we will focus on an interesting observation: the Hessian of the loss function can also be efficiently computed using a new auxiliary linear MHE system. Remarkably, this approach leverages the same Kalman filter and much of the computation that was already used to obtain the gradient.

\subsection{MHE Hessian}
\label{subsec:kalman hessian}
Let ${\rm H}_{\hat{\mathbf{x}}}L$ denote the Hessian of $L$ w.r.t $\hat{\mathbf{x}}$, ${\rm H}_{\boldsymbol \theta}{\hat{\mathbf{x}}}:=\partial {vec}\left (\nabla_{\theta }\hat{\mathbf x} \right )/\partial {\boldsymbol \theta }$ the MHE Hessian, and $\otimes  $ the Kronecker product. The total second-order derivative of $L\left( {\hat {\mathbf{x}}\left(\boldsymbol \varpi \right)} \right)$ w.r.t $\boldsymbol \varpi$ can be obtained using the chain rule of matrix calculus:
\begin{equation}
    \begin{aligned}
        { H}_{\boldsymbol \varpi}L\left( {\hat {\mathbf{x}}\left(\boldsymbol \varpi \right)} \right) &=  \left (\nabla_{\boldsymbol \theta}{\hat{\mathbf{x}}}  \right )^{T}  {\rm H}_{\hat{\mathbf{x}}}L \nabla_{\boldsymbol \theta}{\hat{\mathbf{x}}} \left (\nabla_{\boldsymbol \varpi}{\boldsymbol \theta}  \right )^{2} \\
        &\quad +  \left ( \nabla_{\boldsymbol \varpi}{\boldsymbol \theta} \right )^{T} {\rm H}_{\boldsymbol \theta}{\hat{\mathbf{x}}}\left [ \left ( \nabla_{\hat{\mathbf{x}}} L\right )^{T}\otimes {\bm I} \right ]\nabla_{\boldsymbol \varpi}{\boldsymbol \theta} \\
        &\quad + \left [ \left ( \nabla_{\hat{\mathbf{x}}} L\nabla_{\boldsymbol \theta}{\hat{\mathbf{x}}} \right )\otimes {\bm I} \right ]{\rm H}_{\boldsymbol \varpi}{\boldsymbol \theta}.
    \end{aligned}
\label{eq: hessian of loss}
\end{equation}
Note that all the first-order derivatives have been obtained in computing the gradient of $L\left( {\hat {\mathbf{x}}\left(\boldsymbol \varpi \right)} \right)$ w.r.t $\boldsymbol \varpi$. Computing the second-order derivatives ${\rm H}_{\hat{\mathbf{x}}}L$ and ${\rm H}_{\boldsymbol \varpi}{\boldsymbol \theta}$ is straightforward, which is explicit differentiation of the loss function and the NN, respectively. The major challenge lies in the computation of ${\rm H}_{\boldsymbol \theta}{\hat{\mathbf{x}}}$. 

Given the implicit dependence of $\hat{\mathbf{x}}$ on $\boldsymbol{\theta}$ through KKT conditions, we are motivated to differentiate the KKT conditions (\ref{eq:kkt}) w.r.t $\boldsymbol{\theta}$ twice. This results in the following second-order differential KKT conditions.
\begin{subequations}
    \begin{align}
    \begin{split}
        {H}_{\boldsymbol \theta}\nabla_{\hat{\bm x}_{t-N|t}}{\cal L} &= { {\bm {\mathring L}}_{t - N}^{xx}} {\rm H}_{\boldsymbol \theta} {\hat{\bm x}_{t-N|t}} - {\bm {\mathring P}}{\rm H}_{\boldsymbol \theta}{\hat{\bm x}_{t-N}}+ {\bm {\mathring L}}_{t - N}^{x\theta } \\
& \quad + {\bm {\mathring L}}_{t - N}^{xw}{\rm H}_{\boldsymbol \theta} {\hat{\bm w}_{t-N|t}} - {{\bm {\mathring {F}}}}_{t - N}^T{\rm H}_{\boldsymbol \theta} {{\boldsymbol \lambda}_{t-N}^*}= {\bm 0},
    \end{split}
    \label{eq:boundary 2nd-order differential KKT}\\
    \begin{split}
        {H}_{\boldsymbol \theta}\nabla_{\hat{\bm x}_{k|t}}{\cal L} &= { {\bm {\mathring {L}}}}_k^{xx}{\rm H}_{\boldsymbol \theta} {\hat{\bm x}_{k|t}} + { {\bm {\mathring{L}}}}_k^{xw }{\rm H}_{\boldsymbol \theta} {\hat{\bm w}_{k|t}} - {{\bm {\mathring {F}}}}_k^T{\rm H}_{\boldsymbol \theta} {{\boldsymbol \lambda}_{k}^*}\\
&\quad + {\rm H}_{\boldsymbol \theta} {{\boldsymbol \lambda}_{k-1}^*} + { {\bm {\mathring{L}}}}_k^{x\theta } = {\bm 0},
    \end{split}
    \label{eq:2nd-order differential KKT for x}\\
    \begin{split}
        {H}_{\boldsymbol \theta}\nabla_{\hat{\bm w}_{k|t}}{\cal L} &= { {\bm {\mathring {L}}}}_k^{w x}{\rm H}_{\boldsymbol \theta} {\hat{\bm x}_{k|t}} + { {\bm {\mathring {L}}}}_k^{w w}{\rm H}_{\boldsymbol \theta} {\hat{\bm w}_{k|t}} \\
&\quad - { {\bm {\mathring {G}}}}_k^T{\rm H}_{\boldsymbol \theta} {{\boldsymbol \lambda}_{k}^*} + { {\bm {\mathring {L}}}}_k^{w \theta } = {\bm 0},
    \end{split}
    \label{eq:2nd-order differential KKT for w}\\
    \begin{split}
        {H}_{\boldsymbol \theta}\nabla_{\boldsymbol\lambda^{*}_k}{\cal L} &= \mathrlap{{\rm H}_{\boldsymbol \theta} {\hat{\bm x}_{k+1|t}} - {{ {\bm {\mathring {F}}}}_k} {\rm H}_{\boldsymbol \theta} {\hat{\bm x}_{k|t}} - { {\bm {\mathring {G}}}}_k {\rm H}_{\boldsymbol \theta} {\hat{\bm w}_{k|t}}}\\
        &\quad + \mathrlap{{ {\bm {\mathring {L}}}}_k^{\lambda \theta} = {\bm 0},}
    \end{split}
    \label{eq:2nd-order differential KKT for lambda}
    \end{align}
\label{eq: 2nd-order differential KKT conditions}%
\end{subequations}
where the coefficient matrices are defined below:
\begin{equation}
    \begin{aligned}
        \bm {\mathring{L}}_{k}^{xx} &={\bm I}\otimes {\bm L}_{k}^{xx}, \ \bm {\mathring{F}}_k ={\bm I}\otimes {\bm F}_k, \ \bm {\mathring{G}}_k = {\bm I} \otimes {\bm G}_k,\\
        \bm {\mathring{L}}_{k}^{xw} & = {\bm I} \otimes {\bm L}_{k}^{xw}, \ \bm {\mathring{L}}_{k}^{ww} = {\bm I} \otimes {\bm L}_{k}^{ww}, \ \bm {\mathring{P}}  ={\bm I}\otimes {\bm P}, \\
        \bm {\mathring {L}}_{t-N}^{x \theta } & = \left ( \nabla_{\hat{\bm x}_{t-N|t}}{\bar {\bm f}}_{0}\right )\hat{\bm X}_{t-N|t}+\left (\nabla_{\boldsymbol{\lambda}_{t-N}^*}{\bar {\bm f}}_{0}\right )\boldsymbol{\Lambda}_{t-N}^* \\ 
        &\quad + \left (\nabla_{\hat{\bm w}_{t-N|t}}{\bar {\bm f}}_{0}\right )\hat{\bm W}_{t-N|t}+\nabla_{\boldsymbol{\theta}}{\bar {\bm f}}_{0},\\
        \bm {\mathring {L}}_{k}^{x \theta } &=\left (\nabla_{\hat{\bm x}_{k|t}}\bar{\bm f}_{k}  \right )\hat{\bm X}_{k|t} +\left (\nabla_{\hat{\bm w}_{k|t}}\bar{\bm f}_{k}  \right )\hat{\bm W}_{k|t} \\
        &\quad + \left (\nabla_{\boldsymbol{\lambda}_k^*}\bar{\bm f}_{k}  \right )\boldsymbol{\Lambda}_{k}^*+\nabla_{\boldsymbol{\theta }}\bar{\bm f}_{k},\\
        \bm {\mathring {L}}_{k}^{w \theta } &=\left (\nabla_{\hat{\bm x}_{k|t}}\bar{\bm g}_{k}  \right )\hat{\bm X}_{k|t} +\left (\nabla_{\hat{\bm w}_{k|t}}\bar{\bm g}_{k}  \right )\hat{\bm W}_{k|t} \\
        &\quad + \left (\nabla_{\boldsymbol{\lambda}_k^*}\bar{\bm g}_{k}  \right )\boldsymbol{\Lambda}_{k}^*+\nabla_{\boldsymbol{\theta }}\bar{\bm g}_{k},\\
        \bm {\mathring{L}}_{k}^{\lambda \theta } &= \left ( \nabla_{{\hat{\bm x}_{k|t}}}\bar{\bm h}_{k} \right )\hat{\bm X}_{k|t} + \left ( \nabla_{{\hat{\bm w}_{k|t}}}\bar{\bm h}_{k} \right )\hat{\bm W}_{k|t},
    \end{aligned}
\label{eq: new coefficient matrices}%
\end{equation}
with ${\bar {\bm f}}_{0}$, $\bar{\bm f}_{k}$, $\bar{\bm g}_{k}$, and $\bar{\bm h}_{k}$ being the left hand sides of \eqref{eq:boundary differential KKT}-\eqref{eq:costate differential KKT}.

It is interesting to notice that (\ref{eq: 2nd-order differential KKT conditions}) and (\ref{eq:differential KKT}) share exactly the same structure. The only exception lies in (\ref{eq:2nd-order differential KKT for lambda}) which has an additional term $\bm {\mathring{L}}_{k}^{\lambda \theta }$, not present in (\ref{eq:costate differential KKT}). This term can be considered as a known input to the linear system. As shown in~\ref{subsec: kalman gradient} and \cite{10313083}, (\ref{eq:differential KKT}) can represent the KKT conditions of the auxiliary MHE optimization problem (\ref{eq: auxiliary MHE}), which is subject to the linear system (\ref{eq:auxiliary dynamics}) constructed from (\ref{eq:costate differential KKT}). Given this observation and concept, we can interpret (\ref{eq: 2nd-order differential KKT conditions}) as the KKT conditions of another auxiliary MHE optimization problem. It is subject to a new linear system constructed from (\ref{eq:2nd-order differential KKT for lambda}):
\begin{equation}
    \bm {\mathring{X}}_{k+1}=\bm {\mathring{F}}_{k}\bm {\mathring{X}}_{k}+\bm {\mathring{G}}_{k}\bm {\mathring{W}}_{k}-\bm {\mathring{L}}_{k}^{\lambda \theta },
    \label{eq:linear system for 2nd-order}
\end{equation}
where $\bm {\mathring{X}}_{k}$ and $\bm {\mathring{W}}_{k}$ denote ${\rm H}_{\boldsymbol \theta} {{\bm x}_{k}}$ and ${\rm H}_{\boldsymbol \theta} {{\bm w}_{k}}$, respectively. The related cost function can be established directly from (\ref{eq:auxiliary mhe cost}) by updating the latter's coefficient matrices, for example, by replacing $\bm P$ with $\bm {\mathring P}$ defined in (\ref{eq: new coefficient matrices}). Consequently, we can efficiently reuse Algorithm~\ref{alg: analytical solution} to solve for the MHE Hessian ${\rm H}_{\boldsymbol \theta} {\hat{\mathbf {x}}}$, which is the optimal solution $\mathbf{\hat{\mathring {{X}}}}$ to the new auxiliary linear MHE problem. Only the algorithm's inputs need minor modifications as follows: 
\begin{itemize}
    \item Replace $\hat {\bm X}_{t-N}$ with $\bm{\hat{\mathring {X}}}_{t-N}$ (which is approximated by ${\rm H}_{\boldsymbol \theta} {\hat{\bm x}_{t-N|t-1}}$ obtained at $t-1$);
    \item Update the matrices in (\ref{eq: kalman matrices}) using (\ref{eq: new coefficient matrices}). In particular, add $\bm {\mathring{L}}_{k}^{\lambda \theta }$ to (\ref{eq: kalman filter prediction}) such that the updated ${\bm A}_k$ becomes ${\bm {\mathring {A}}_k} = {\bm {\mathring {G}}_{k}}{\left( { {\bm {\mathring {L}}}_{k}^{w w }} \right)^{ - 1}}{\bm {\mathring {L}}}_{k}^{w \theta } + \bm {\mathring{L}}_{k}^{\lambda \theta }$. This can be derived based on the example used to obtain (\ref{eq: kalman filter prediction}) (See \ref{subsec: kalman solver}).
\end{itemize}

\subsection{Kalman Filter-based Solver}
\label{subsec: kalman solver}
First, a Kalman filter (KF) is solved to obtain the estimates $\left\{ {\hat {\bm X}_{k\left| k \right.}^{\rm KF}}\right\} _{k = t-N}^t$: the initial condition is given by
\begin{equation}
     {\hat {\bm X}_{t - N\left| t \right. - N}^{\rm KF}} =\left( {\bm I + {\bm C_{t - N}}{\bm S_{t - N}}} \right)\hat {\bm X}_{t-N}  + {\bm C_{t - N}}{\bm T_{t - N}}.  
\label{eq: initial condition for x}
\end{equation}
where $\bm I$ is an identity matrix, ${\bm C_{t - N}}$ is defined in (\ref{eq: kalman filter covariance update}) with ${\bm P_{t-N}} = {\bm P^{-1}}$, ${\bm S_{t - N}}$ and ${\bm T_{t - N}}$ are given in (\ref{eq: kalman matrices}), and $\hat {\bm X}_{t-N}$ is approximated by $\nabla_{\boldsymbol \theta} {\hat{\bm x}_{t-N|t-1}}$ obtained at $t-1$. Then, the remaining estimates are generated through the following equations from $k=t-N+1$ to $t$:
\begin{subequations}
   \begin{align}
       {{\hat {\bm X}}_{k\left| {k - 1} \right.}} & = {{\bar {\bm F}}_{k - 1}}{{\hat {\bm X}}_{k - 1\left| {k - 1} \right.}^{\rm KF}} - {\bm A_{k-1}},
       \label{eq: kalman filter prediction}\\
       {\bm P_k} & = {{\bar {\bm F}}_{k - 1}}{\bm C_{k - 1}}\bar {\bm F}_{k - 1}^T + {\bm B_{k-1}},
       \label{eq: kalman filter covariance}\\
       {\bm C_k}& = {\left( {\bm I - {\bm P_k}{\bm S_k}} \right)^{ - 1}}{\bm P_k},
       \label{eq: kalman filter covariance update}\\
       {{\hat {\bm X}}_{k\left| k \right.}^{\rm KF}} & = \left( {\bm I + {\bm C_k}{\bm S_k}} \right){{\hat {\bm X}}_{k\left| {k - 1} \right.}} + {\bm C_k}{\bm T_k}.
       \label{eq: kalman filter estimate}
   \end{align}
    \label{eq: kalman}%
\end{subequations}
where ${\bm A_k}$, ${\bm B_k}$, ${\bm S_k}$, ${\bm T_k}$, and ${\bar {\bm F}_k}$ are defined below:
\begin{equation}
\begin{aligned}
    {\bm A_k}  & = {\bm G_{k }}{\left( { {\bm L}_{k }^{w w }} \right)^{ - 1}}{\bm L}_{k }^{w \theta },\\
    {\bm B_k}  & = {\bm G_{k }}{\left( { {\bm L}_{k }^{w w }} \right)^{ - 1}}\bm G_{k }^T,\\
    {\bm S_k}  & = {\bm L}_k^{xw }{\left( { {\bm L}_k^{w w }} \right)^{ - 1}}{\bm L}_k^{w x} - \bar {\bm L}_k^{xx}, \ {\bm S_t}  = - \bar {\bm L}_t^{xx},\\
    {\bm T_k}  & = {\bm L}_k^{xw }{\left( { {\bm L}_k^{w w }} \right)^{ - 1}}{\bm L}_k^{w \theta } - {\bm L}_k^{x\theta }, \ {\bm T_t}  = - {\bm L}_t^{x\theta },\\
    {\bar {\bm F}_k}  & = {\bm F_k} - {\bm G_k}{\left( {{\bm L}_k^{w w }} \right)^{ - 1}} {\bm L}_k^{w x}.
\label{eq: kalman matrices}
\end{aligned}
\end{equation}

Second, the new dual variables $\bm \Lambda^*  = \left\{ {{\bm \Lambda^* _k}} \right\}_{k = t - N}^{t - 1}$ are computed using the following equation backward in time from $k=t$ to $t-N+1$, starting with $\bm \Lambda^*_t = \bm 0$:
\begin{equation}
    {\bm \Lambda^* _{k - 1}} = \left( {\bm I + {\bm S_k}{\bm C_k}} \right)\bar {\bm F}_k^T{\bm \Lambda^* _k} + {\bm S_k}{\hat {\bm X}_{k\left| k \right.}^{\rm KF}} + {\bm T_k}.
    \label{eq: backward lambda}
\end{equation}

Third, the optimal estimates $\hat{\bm {\mathbf{X}}}$ are obtained using 
\begin{equation}
    {\hat {\bm X}_{k\left| t \right.}} = {\hat {\bm X}_{k\left| k \right.}^{\rm KF}} + {\bm C_k}\bar {\bm F}_k^T{\bm \Lambda^* _k},
    \label{eq: forward state}
\end{equation}
iteratively from $k=t-N$ to $t$.

The above recursions are summarized in Algorithm~\ref{alg: analytical solution}.

\begin{algorithm}[!h]
\caption{Solving for $\hat {\mathbf{X}}$ using a Kalman filter}
\label{alg: analytical solution}
\SetKwInput{Input}{Input}
\SetKwInput{Output}{Output}
\SetKw{by}{by}
\SetKwProg{Pn}{def}{:}{\KwRet $\left\{ {{{\hat {\bm X}}_{k\left| t \right.}}} \right\}_{k = t - N}^t$}
\Input{$\hat {\bm X}_{t-N}$ and the matrices in (\ref{eq: kalman matrices});}
\Indp \Pn{Kalman\_Filter\_based\_Gradient\_Solver}{
Set ${{\hat {\bm X}}_{{t-N}\left| {{t-N}} \right.}^{\rm KF}}$ using (\ref{eq: initial condition for x});\\
\For{$k \leftarrow t-N+1$ \KwTo $t$ \by $1$}{
Update ${{\hat {\bm X}}_{{k}\left| {{k}} \right.}^{\rm KF}}$ using the Kalman filter (\ref{eq: kalman});
}
\For{$k \leftarrow t$ \KwTo $t-N+1$ \by $-1$}{
Update ${{\bm \Lambda^*_{k-1}} }$ using (\ref{eq: backward lambda}) with ${{\bm \Lambda^*_t}} = {\bm 0}$;
}
\For{$k \leftarrow t-N$ \KwTo $t$ \by $1$}{
Update ${\hat {\bm X}_{{k}\left| t \right.}}$ using (\ref{eq: forward state}); 
}
}
\Indm\Output{$\nabla_{\boldsymbol \theta} {\hat{\mathbf {x}}}= {\hat {\bm {\mathop{ {\rm X}}\nolimits}}}$ }
\end{algorithm}

In the Kalman filter (\ref{eq: kalman}), (\ref{eq: kalman filter prediction}) is the state predictor, (\ref{eq: kalman filter covariance}) handles the error covariance prediction, (\ref{eq: kalman filter covariance update}) addresses the error covariance correction, and (\ref{eq: kalman filter estimate}) is the state corrector. These equations are derived from the differential KKT conditions (\ref{eq:differential KKT}) by induction. For example, we obtain (\ref{eq: kalman filter prediction}) by following steps: 1) solve for  $\hat{\bm W}_{k|t}:=\nabla_{\boldsymbol \theta} {\hat{\bm w}_{k|t}}$ from (\ref{eq:w differential KKT}); 2) plug $\hat{\bm W}_{k|t}$ and $\hat{\bm X}_{k|t}$ (obtained from (\ref{eq: forward state})) into (\ref{eq:costate differential KKT}); 3) define the resulting equation, excluding the term related to $\boldsymbol\Lambda^*_k:=\nabla_{\boldsymbol \theta} {{\boldsymbol \lambda}_{k}^*}$, as $\hat{\bm X}_{k+1|k}$ using $\bm A_k$ and $\bar{\bm F}_k$ given in (\ref{eq: kalman matrices}). The detailed theoretical justification of (\ref{eq: kalman}) can be found in Appendix-C of~\cite{10313083}. 
Note that (\ref{eq: kalman}) is not presented in the standard form of a Kalman filter for the purpose of a clearer and more inductive proof, as explained in \cite{10313083}. The Kalman filter provides analytical gradients in a recursive form, significantly enhancing computational efficiency.

\section{Trust-Region Learning Framework}\label{sec:Trust Region}
We aim to train the neural network's parameters $\boldsymbol \varpi$ using the trust-region method, which addresses Problem (\ref{eq:bi level neuromhe}) by creating a trust region around the current parameters $\boldsymbol \varpi_t$. Within this region, it minimizes a locally quadratic approximation of $L\left( {\hat {\mathbf{x}}\left(\boldsymbol \varpi_t \right)} \right)$ to determine a candidate update $\boldsymbol{\mu}$ for $\boldsymbol\varpi _{t}$. The corresponding optimization problem is as follows:
\begin{subequations}
    \begin{align}
        &\mathop {\min }\limits_{\boldsymbol\mu } {\textsc{TRM} }\left (\boldsymbol\mu \right ):=L+ \boldsymbol\mu^{T}\nabla_{\boldsymbol\varpi_t}L + \frac{1}{2}\boldsymbol\mu^{T}\left ({H}_{\boldsymbol\varpi_t}L  \right )\boldsymbol\mu\\
        &\ {\rm s.t.}\ \left \|  \boldsymbol\mu   \right \| \leq \Delta _{t}.
        \label{eq:trust-region constraint}
    \end{align}
\label{eq: trust-region problem}%
\end{subequations}
Given predefined thresholds $0\leq \xi _{1}\leq \xi _{2}< \xi _{3}$, $\kappa _{1}\in \left ( 0,1 \right )$, $\kappa _{2}> 1$, and the upper bound of the radius $\bar\Delta$, the method iteratively explores the range of trust region by updating its radius $\Delta _{t}\in \mathbb{R}_{+}$ with the following law~\cite{nocedal1999numerical}:
\begin{equation}
\Delta _{t+1}=\left\{\begin{matrix*}[l]
\kappa _{1}\Delta _{t} & {\rm if}\ \rho _{t}< \xi _{2}\\ 
\min\left ( \kappa _{2}\Delta _{t},\bar\Delta \right ) & {\rm if}\ \rho _{t}> \xi _{3} \&  \left \| \boldsymbol\mu _{t} \right \|= \Delta _{t} \\ 
\Delta _{t} & {\rm otherwise}
\end{matrix*}\right.
\label{eq: radius updating}
\end{equation}
Here, $\rho_t$ denotes the current ratio between the actual and predicted reductions on the loss function
\begin{equation}
    \rho _{t}=\frac{L\left ( \hat{\mathbf x}\left ( \boldsymbol \varpi _{t} \right ) \right )-L\left ( \hat{\mathbf x}\left ( \boldsymbol \varpi _{t} + \boldsymbol \mu _{t}\right ) \right )}{{\textsc{TRM}\left ( \mathbf{0} \right )-\textsc{TRM}\left ( \boldsymbol \mu _{t} \right )}},
    \label{eq:ratio for trust-region}
\end{equation}
which is used to decide whether to update $\boldsymbol \varpi_t$: if $\rho _{t}> \xi _{1}$, $\boldsymbol \varpi_t$ will be updated using $\boldsymbol\mu _{t}$. We summarize the trust-region policy optimization for training NeuroMHE in Algorithm~\ref{alg: trust-region}.

\begin{algorithm}[!h]
\caption{Trust-region NeuroMHE training}
\label{alg: trust-region}
\SetKwInput{Input}{Input}
\SetKwInput{Output}{Output}
\SetKw{by}{by}
\Input{The desired states $\cal S$ and the measurements $\cal Y$.}

\While {$L_{\rm mean}$ not converged}{
\For{$t \leftarrow 0$ \KwTo $T$}
{Obtain ${\hat {\mathbf{x}}\left(\boldsymbol \varpi_t \right)}$ by solving $\rm{NeuroMHE}\left(\boldsymbol \varpi_t \right)$ with the most recent measurements $\mathbf{y}_t\in \cal {Y}$;\\
Compute $L\left ( \hat{\mathbf x}\left ( \boldsymbol \varpi _{t} \right ) \right )$ using ${\hat {\mathbf{x}}\left(\boldsymbol \varpi_t \right)}$ and $\mathbf{s}\in \cal {S}$;\\
Compute $\nabla_{\boldsymbol\varpi_t}L$ using (\ref{eq: gradient of loss}) and Algorithm~\ref{alg: analytical solution};\\
Compute ${H}_{\boldsymbol\varpi_t}L$ using (\ref{eq: hessian of loss}) and Algorithm~\ref{alg: analytical solution} with its inputs updated by (\ref{eq: new coefficient matrices});\\
Obtain $\boldsymbol\mu _{t}$ by solving $\textsc{TRM}$ Problem (\ref{eq: trust-region problem});\\
Obtain ${\hat {\mathbf{x}}\left(\boldsymbol \varpi_t + \boldsymbol\mu _{t} \right)}$ by solving $\rm{NeuroMHE}\left(\boldsymbol \varpi_t+ \boldsymbol\mu _{t}\right)$ with the same $\mathbf{y}_t$;\\
Compute $L\left ( \hat{\mathbf x}\left ( \boldsymbol \varpi _{t} + \boldsymbol \mu _{t}\right ) \right )$ using $\hat{\mathbf x}\left ( \boldsymbol \varpi _{t} + \boldsymbol \mu _{t}\right )$ and the same $\mathbf{s}$;\\
Update $\rho _{t}$ using (\ref{eq:ratio for trust-region});\\
Update $\Delta_t$ using (\ref{eq: radius updating});\\
\uIf{$\rho _{t}> \xi _{1}$}{Update $\boldsymbol\varpi _{t+1}\leftarrow \boldsymbol\varpi _{t} + \boldsymbol\mu _{t}$;}
\Else{$\boldsymbol\varpi _{t+1}\leftarrow \boldsymbol\varpi _{t}$;}
}
{Compute the mean loss $L_{\rm mean}=\frac{1}{T}\sum_{t=0}^{T} L\left ( \hat{\mathbf{x}}\right )$ for the next episode}\\
}

\end{algorithm}

\section{Numerical Case Study}\label{sec:experiment}

We numerically validate the performance of trust-region NeuroMHE in estimating complex aerodynamic disturbances on quadrotors using a real flight dataset from various agile flights~\cite{bauersfeld2021neurobem}. We compare it with the gradient descent counterpart and the state-of-the-art estimator NeuroBEM~\cite{bauersfeld2021neurobem} to showcase the following advantages: 1) highly-efficient training; 2) improved robustness to network initialization; and 3) superior force estimation accuracy.

In alignment with NeuroBEM, NeuroMHE is trained using estimation errors. In Algorithm~\ref{alg: trust-region}, we define the desired states $\cal S$ using the ground truth disturbance forces $\bm F$ and torques $\boldsymbol{\tau}$ from the real dataset and the quadrotor model. For NeuroMHE, the measurements $\cal Y$ are set as the quadrotor's linear $\bm v$ and angular velocities $\boldsymbol \omega$. By comparison, the NeuroBEM's inputs also comprise motor speeds, requiring specific sensors and autopilot firmware. To address the state estimation problem (\ref{eq:mhe}), we integrate the velocities with the unmeasurable disturbances to create the augmented state ${\bm x}=\left [ {\bm v};{\bm F};\boldsymbol\omega ;\boldsymbol\tau  \right ]\in \mathbb{R}^{12}$. Assuming no loss of generality, the disturbance dynamics are driven by the process noises ${\bm w}_f \in \mathbb{R}^3$ and ${\bm w}_{\tau} \in \mathbb{R}^3$. The system model for state prediction in MHE is as follows:
\begin{subequations}
\begin{align}
        \dot {\bm v} & = {m^{ - 1}}\left( -mg {\bm e}_3  + {\bm F} \right),& {\dot {\bm F}} & = {\bm w _f}
        \label{eq:position reduced dynamics},\\
        \dot {\boldsymbol \omega}  & = {{\bm J}^{ - 1}}\left(  - {\bm \omega ^ \times } {\bm J}{\boldsymbol \omega}  + {{\boldsymbol \tau}} \right),& {\dot {\boldsymbol{\tau}}} & = {\bm w _{\tau}},
        \label{eq:attitude reduced dynamics}
\end{align}
\label{eq:quadrotor reduced model}%
\end{subequations}
where $g=9.81\ {\rm m}/{\rm s^2}$, ${\bm e}_{3}=\left [ 0;0;1 \right ]$, $m = 0.772\ {\rm kg}$, and ${\bm{J}} = {\rm{diag}}\left( {0.0025,0.0021,0.0043} \right)\ {\rm{kg}} {{\rm{m}}^{\rm{2}}}$~\footnote{These inertial values are reported at the NeuroBEM's website~\url{https://rpg.ifi.uzh.ch/neuro_bem/Readme.html}.}.

The weighting matrices of NeuroMHE have the following dimensions: ${\bm P}\in\mathbb{R}^{12\times 12}$, ${\bm R}_k\in\mathbb{R}^{6\times 6}$, and ${\bm Q}_k\in\mathbb{R}^{6\times 6}$. Instead of training the neural network to generate all ${\bm R}_k$ and ${\bm Q}_k$ over an MHE horizon, we introduce two forgetting factors ${\gamma _{1,2}} \in \left( {0,1} \right)$ to parameterize the time-varying ${\bm R}_k$ and ${\bm Q}_k$ by ${{\bm R}_k} = \gamma _1^{t - k}{{\bm R}_t}$ and ${{\bm Q}_k} = \gamma _2^{t - 1 - k}{{\bm Q}_{t - 1}}$. To minimize the network's size, we configure ${\bm P}$, ${\bm R}_t$, and ${\bm Q}_{t - 1}$ as diagonal matrices. Finally, the diagonal elements are specified as ${P_{\cdot}} = \varsigma  + p_{\cdot}^2$, ${R_{\cdot}} = \varsigma  + r_{\cdot}^2$, and ${Q_{\cdot}} = \varsigma  + q_{\cdot}^2$, where $\varsigma  > 0$, to guarantee the positive definiteness. The gradient (\ref{eq: gradient of loss}) and Hessian (\ref{eq: hessian of loss}) are updated accordingly to incorporate the above parameterization. Our neural network has two hidden layers with the Leaky-ReLU activation function. It takes the measurement ${\bm y}=\left [ {\bm v};\boldsymbol\omega  \right ]$ as inputs, has $8$ neurons in each hidden layer, and outputs $\boldsymbol\Theta = \left [ p_{1:12},{\gamma}_1,r_{1:6},{\gamma}_2,q_{1:6} \right ]\in {\mathbb R}^{26}$, resulting in a total of $362$ network parameters. 

\begin{figure}[h]
\centering
\begin{subfigure}[b]{0.49\textwidth}
\centering
{\includegraphics[width=0.925\textwidth]{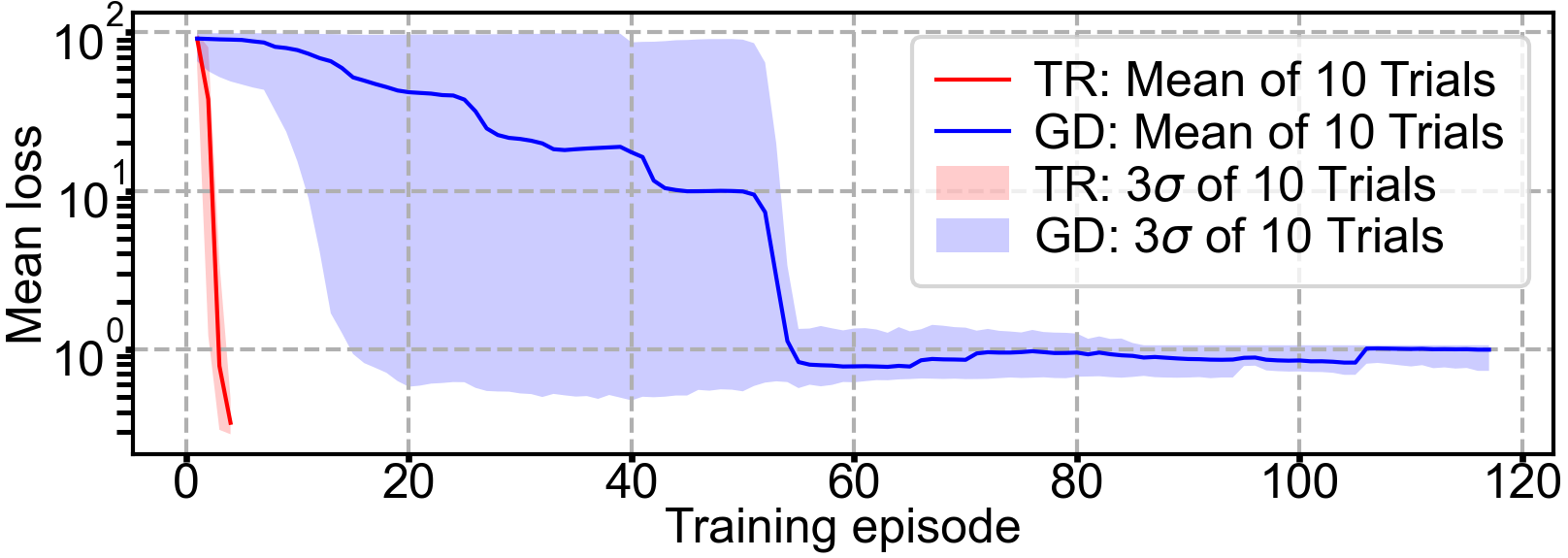}}
\caption{\footnotesize Mean loss in 10 trials.}
\label{fig:meanloss comparison}
\end{subfigure}
\hfill
\begin{subfigure}[b]{0.235\textwidth}
\centering
\includegraphics[width=0.85\textwidth]{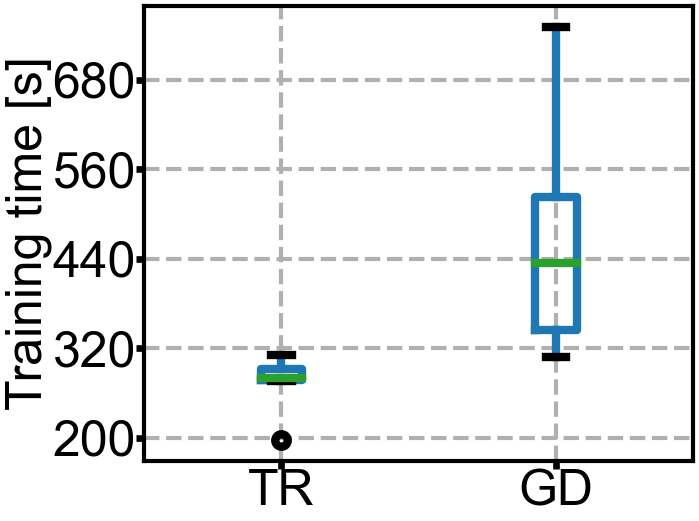}
\caption{\footnotesize Overall training time.}
\label{fig:training time}
\end{subfigure}
\hfill
\begin{subfigure}[b]{0.235\textwidth}
\centering
\includegraphics[width=0.85\textwidth]{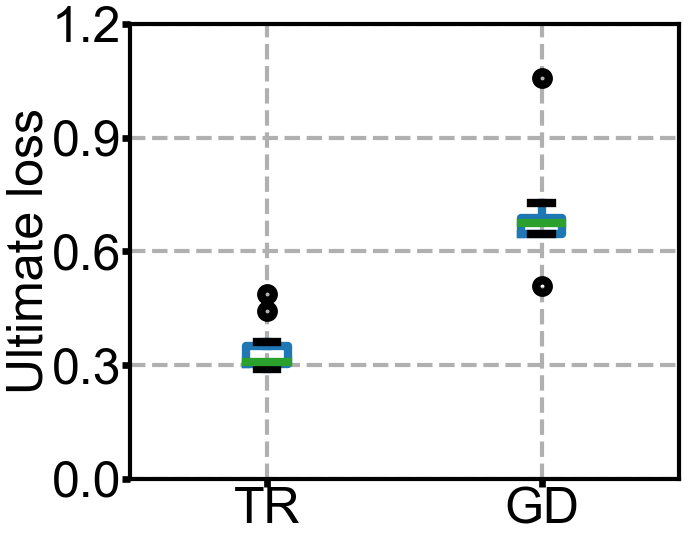}
\caption{\footnotesize Ultimate loss.}
\label{fig:steady mean loss}
\end{subfigure}
\caption{\footnotesize Comparison of the training performance between the gradient descent (GD) and the trust-region (TR) methods. We randomly initialize the neural network in $10$ trials with the Kaiming initialization, a common method for ReLU~\cite{he2015delving}. For the GD method, we set the learning rate to $1\times 10^{-4}$ by balancing between training stability and performance. We carefully select the trials of gradient descent such that the untrained mean loss closely matches that of the TR method, enabling fair comparisons. The ultimate loss shown in Fig.~\ref{fig:steady mean loss} corresponds to the loss value in the last episode.}
\label{fig: training comparison}
\end{figure}

For training, we select a $0.25$-s-long trajectory segment from a Figure-8 flight collected at 400Hz (a total of $100$ data points). One training episode amounts to training NeuroMHE over this $0.25$-s-long dataset once. For testing, we adopt the same dataset as in~\cite{bauersfeld2021neurobem} to compare NeuroMHE with NeuroBEM across $13$ unseen agile trajectories. We implement our algorithm using CasADi~\cite{andersson2019casadi} and conduct the simulations in a workstation with an Intel Core i7-11700K processor.

\begin{table}[h]
\fontsize{8}{8}\selectfont
\caption{Runtime of Algorithm~\ref{alg: analytical solution} for Different MHE Horizons\label{table:cpu time}}
\centering
\begin{threeparttable}[h]
\begin{tabular}{ c|c c c c c } 
\toprule[1pt]
Horizon $N$ & $10$ & $20$ & $40$ & $60$ & $80$ \\
\midrule[0.5pt]
MHE Hessian $[{\rm ms}]$ & $67.5$ & $137.8$ & $253.4$ & $379.3$ & $516.1$\\
MHE Gradient $[{\rm ms}]$& $1.83$ & $3.74$ & $6.97$ & $12.36$ & $14.93$\\
\bottomrule[0.5pt]
\end{tabular}
\end{threeparttable}
\end{table}

Before showing the estimation performance of trust-region NeuroMHE, we assess its training efficiency by comparing the CPU runtime of Algorithm~\ref{alg: analytical solution} and the overall training time with gradient descent. Table~\ref{table:cpu time} shows that both methods exhibit roughly linear computational complexity w.r.t the horizon, indicating scalability for large MHE problems. Computing the MHE Hessian takes longer than the MHE gradient due to the $\otimes$ operation, which expands the matrix dimension (See (\ref{eq: new coefficient matrices})). Despite this, the trust-region method significantly reduces the overall training time (under $5\ {\rm min}$) by up to $63\%$ (See Fig.~\ref{fig:training time}) while producing better training results (See Fig.~\ref{fig:steady mean loss}). Given the quadrotor's fast dynamics in agile flights, we set $N=10$ in the subsequent tests.

\begin{figure}[h]
\centering
\begin{subfigure}[b]{0.238\textwidth}
\centering
\includegraphics[width=0.985\textwidth]{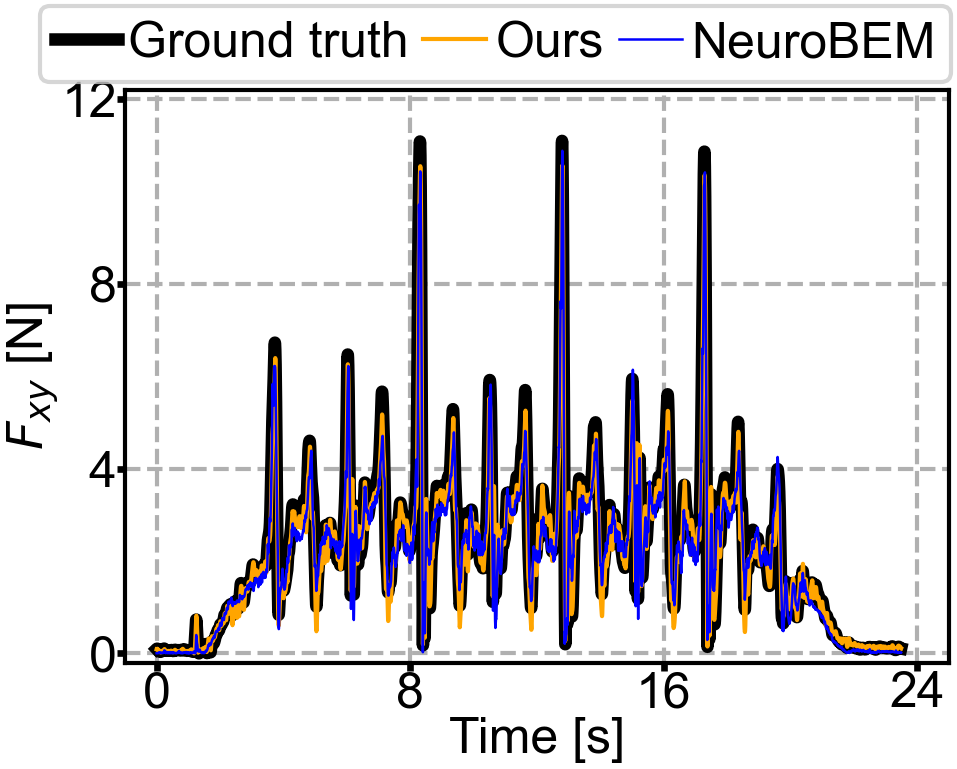}
\caption{\footnotesize Estimation of $F_{xy}$.}
\label{fig:fxy}
\end{subfigure}
\hfill
\begin{subfigure}[b]{0.238\textwidth}
\centering
\includegraphics[width=0.985\textwidth]{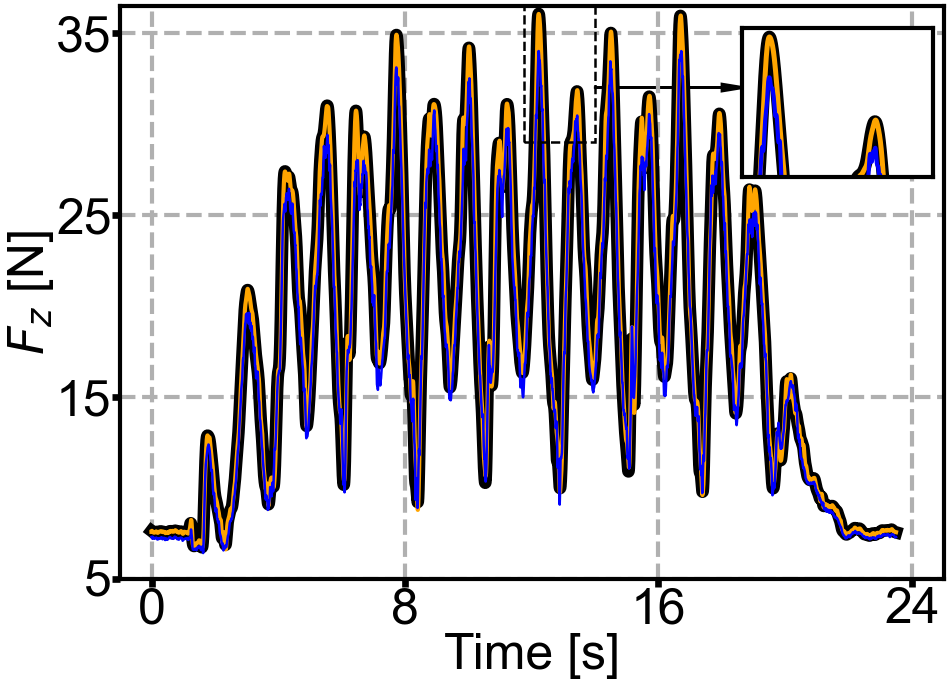}
\caption{\footnotesize Estimation of $F_{z}$.}
\label{fig:fz}
\end{subfigure}
\caption{\footnotesize Comparison of the force estimation performance between NeuroMHE and NeuroBEM on an aggressive Figure-8 flight test dataset as used in~\cite{bauersfeld2021neurobem}.}
\label{fig: force estimation}
\end{figure}

\begin{table}[h]
\fontsize{8}{8}\selectfont
\caption{RMSE Comparisons on the Figure-8 Flight Test Dataset\label{table:rmse}}
\centering
\begin{threeparttable}[t]
\begin{tabular}{ c|c c c c| c c } 
\toprule[1pt]
\multirow{2}{*}{Method}  & ${F_{xy}}$ & ${F_z}$ & ${\tau_{xy}}$ & ${\tau_{z}}$ & $F$ & ${\tau}$ \\
       & $\left[ {\rm N} \right]$ & $\left[ {\rm N} \right]$ & $\left[ {\rm Nm} \right]$ & $\left[ {\rm Nm} \right]$& $\left[ {\rm N} \right]$ & $\left[ {\rm Nm} \right]$ \\
\midrule[0.5pt]
{ NeuroBEM}  & $0.51$ &  $1.08$ & $0.03$ & $0.01$& $1.20$ & $0.03$ \\
{ Ours}  & ${\bf 0.32}$ & ${\bf 0.31}$ &${\bf 0.02}$ & ${ 0.01}$&${\bf 0.45}$&${\bf 0.02}$ \\
\bottomrule[0.5pt]
\end{tabular}
\end{threeparttable}
\end{table}

Fig.~\ref{fig: force estimation} compares the performance of trust-region NeuroMHE with NeuroBEM in estimating force on an unseen flight test dataset. Our method maintains high accuracy, even in challenging force spikes where NeuroBEM experiences substantial performance degradation. Table~\ref{table:rmse} presents the related quantitative comparison using Root-Mean-Square errors (RMSEs). Our approach significantly outperforms NeuroBEM in the planar and vertical force estimations, reducing the RMSEs by $37.3\%$ and $71.3\%$, respectively. Further comparisons on the entire test dataset\footnote{The comparison results on the entire $13$ unseen test flight trajectories are available at \url{https://github.com/BinghengNUS/TR-NeuroMHE}.} show that our method substantially improves the overall force estimation by up to $68.1\%$ across most trajectories. Note that this superior generalizability is achieved by training a lightweight network with $362$ parameters on just $0.25$-s-long data. By comparison, NeuroBEM requires $3150$-s-long flight data to train a high-capacity neural network with $25{\rm k}$ parameters.

\section{Conclusion and Future Work}\label{sec:conclusion} 
This paper proposed a trust-region policy optimization method for training NeuroMHE. Our critical insight is that most of the computation initially used to obtain the MHE gradient, especially the Kalman filter, can be efficiently reused for calculating the MHE Hessian in a recursive form. Through extensive simulations using real flight data, we have shown that our method provides highly-efficient training, accurate estimation with fast online adaptation to various challenging scenarios, and improved robustness to network initialization. 

Our future work includes conducting a theoretical analysis on the stability of trust-region NeuroMHE, testing its efficacy in solving large scale estimation problems with various activation functions in the neural network, and simultaneously learning optimal controllers and estimators using second-order optimization techniques.

\addtolength{\textheight}{-12cm}   





\bibliographystyle{IEEEtran}
\addtolength{\textheight}{5cm}
\bibliography{reference}

\end{document}